\pdfoutput=1
\documentclass{article}

\PassOptionsToPackage{numbers,compress}{natbib}
\usepackage[main, final]{neurips_2026}

\usepackage[utf8]{inputenc}
\usepackage[T1]{fontenc}
\usepackage{hyperref}
\usepackage{url}
\usepackage{booktabs}
\usepackage{amsmath}
\usepackage{amssymb}
\usepackage{amsthm}
\usepackage{amsfonts}
\usepackage{microtype}
\usepackage[table]{xcolor}
\usepackage{multirow}
\usepackage{makecell}
\usepackage{threeparttable}
\usepackage{float}
\usepackage{algorithm}
\usepackage{algorithmic}
\usepackage{graphicx}
\usepackage{subcaption}
\usepackage{wrapfig}
\usepackage{siunitx}
\usepackage{array}
\usepackage{enumitem}

\newcommand{\levelOne}{\ensuremath{\mathrm{L}_{1}}}
\newcommand{\levelTwo}{\ensuremath{\mathrm{L}_{2}}}
\newcommand{\levelThree}{\ensuremath{\mathrm{L}_{3}}}
\newcommand{\levelFour}{\ensuremath{\mathrm{L}_{4}}}

\newcommand{\Rhat}{\widehat{\mathcal{R}}_{\mathrm{FA}}}
\newcommand{\RhatZero}{\widehat{\mathcal{R}}^{\,0}_{\mathrm{FA}}}
\newcommand{\RhatPlus}{\widehat{\mathcal{R}}^{+}_{\mathrm{FA}}}

\newcommand{\dpos}[1]{\,{\scriptsize\textcolor{gray}{(+#1)}}}
\newcommand{\dneg}[1]{\,{\scriptsize\textcolor{gray}{($-$#1)}}}
\newcommand{\dposp}[1]{\,{\scriptsize\textcolor{gray}{(\phantom{0}+#1)}}}
\newcommand{\dnegp}[1]{\,{\scriptsize\textcolor{gray}{(\phantom{0}$-$#1)}}}
\definecolor{AuditSummaryBg}{gray}{0.92}
\DeclareRobustCommand{\auditbglabel}[1]{\begingroup\setlength{\fboxsep}{1pt}\colorbox{AuditSummaryBg}{#1}\endgroup}
\newcommand{\Judge}{\mathcal{J}}
\newcommand{\Aset}{\mathcal{A}}
\newcommand{\Bpix}{\mathcal{B}_{\text{pix}}}
\newcommand{\Btxt}{\mathcal{B}_{\text{txt}}}
\newcommand{\Bk}{\mathcal{B}_{k}}
\newcommand{\phik}{\phi_{k}}
\newcommand{\phipix}{\phi_{\text{pix}}}
\newcommand{\phitxt}{\phi_{\text{txt}}}
\newcommand{\Set}[1]{\mathcal{S}_{#1}}

\title{FusionAudit: Pathway-Conditioned Robustness Auditing for Native Multimodal Models}

\author{Jiaming Zhang \qquad Zherui Li \qquad Siqi Guo \\
  \textbf{Junhao Dong \qquad Kun Wang \qquad Wei Yang Bryan Lim} \\[4pt]
  College of Computing and Data Science \\ Nanyang Technological University, Singapore \\
  Project page: \url{https://jiamingzhang.netlify.app/fusionaudit}
}

\begin{document}

\maketitle

\begin{abstract}
As vision-language models (VLMs) transition from frozen-encoder pipelines to native multimodal architectures, the visual channel increasingly serves as both a perceptual input and a carrier of rendered text that competes with user prompts. Current single-attack, pixel-centric evaluations fail to capture this complexity: they collapse distinct visual attack surfaces into a single scalar and ignore how architectural shifts alter vulnerability. We introduce \textbf{PaRMA}, an evidence-based auditing framework that defines pathway-conditioned risk over two primary image-level interventions: pixel perturbation and rendered-text insertion. Instead of arbitrary attack enumeration, PaRMA uses pathway-specific diagnostics to expose architecture-specific blind spots. Notably, we uncover a \textit{see-but-not-deceived} signature where native VLMs attend heavily to typographic overlays but visually reject them, and we identify an objective mismatch between encoder-level and answer-level attacks in deep-fusion models. To exploit these findings, we instantiate \textbf{FusionAudit}, introducing two diagnostic-motivated attacks: \textbf{MANTIS} (margin-aware pixel perturbation) and \textbf{Nat-Evid} (host-congruent semantic insertion). Evaluating 18 open-weight VLMs across a novel nativeness taxonomy, we demonstrate that legacy single-attack rankings (e.g., PGD) are unstable for native models. Our diagnostic-driven attacks substantially raise the estimated residual risk, showing that PaRMA provides a falsifiable, evidence-driven auditing methodology for uncovering vulnerabilities missed by standard benchmarks.
\end{abstract}

\section{Introduction}
\label{sec:intro}

Vision-language models (VLMs) are undergoing a structural transition. Early systems treated the visual channel as a peripheral sensor, piping a frozen vision encoder into a pretrained language model via a projection bottleneck. Today, native multimodal architectures---featuring multi-layer visual injection and early-fusion shared backbones~\cite{qwen3vl, team2024chameleon, gemma4_2026}---process visual and textual tokens in a unified representational space from the earliest layers. In this paradigm, pixels, scene text, and document fields compete directly with the user's prompt. However, robustness evaluation has not kept pace. Current protocols remain siloed into pixel-space adversarial attacks~\cite{zhao2024evaluating, carlini2024aligned} and typographic overlays~\cite{goh2021multimodal}, neither of which accounts for the deep integration of native VLMs.

This structural gap leaves two critical blind spots in current evaluations. First, on the pixel pathway, legacy attacks anchored at the vision encoder ignore deep decoder-side gradients. Conversely, standard end-to-end attacks (e.g., PGD~\cite{madry2018towards}) spread gradient pressure across all output tokens, underweighting the exact position where the model commits to a wrong answer. Second, on the rendered-text pathway, native VLMs exhibit a striking \textit{see-but-not-deceived} signature: attention diagnostics indicate that these models allocate heavy attention to overlaid text, yet they resist adopting it as a task response. This suggests native VLMs develop a fine-grained sensitivity to visual congruence, making direct typographic attacks incomplete and leaving additional semantic failures unmeasured due to visual rejection.

We argue that robustness auditing for native VLMs should be an evidence-based methodology rather than a static benchmark. We introduce \textbf{PaRMA} (Pathway-conditioned Risk for Multimodal Auditing), a framework that organizes the visual channel into two primary intervention families: pixel perturbation ($\pi_{\mathrm{pix}}$) and rendered-text insertion ($\pi_{\mathrm{txt}}$). Instead of blindly enumerating attacks, PaRMA uses behavioral and gradient diagnostics to expose architecture-specific blind spots, which then motivate the design of diagnostic-motivated interventions. We instantiate this framework as \textbf{FusionAudit}. Guided by our diagnostic findings, FusionAudit introduces two strengthened attacks: \textbf{MANTIS}, a margin-aware pixel attack that concentrates gradients at the first-divergence token to address this mismatch; and \textbf{Nat-Evid}, a host-congruent semantic intervention that naturalizes text payloads to bypass visual rejection. The audit's principal scalar is the residual pathway-complete risk, yielding an empirical lower bound on pathway-complete risk.

\textbf{Contributions.} Our core contributions are:
\begin{itemize}[leftmargin=*, nosep]
    \item We identify native VLM robustness as a structurally distinct problem, introducing a graded \textit{nativeness taxonomy} (\levelOne{}--\levelThree{}) to organize audit targets across fusion designs.
    \item We propose PaRMA and FusionAudit, an evidence-based auditing methodology. We uncover the \textit{see-but-not-deceived} signature in native VLMs and introduce diagnostic-motivated attacks (MANTIS, Nat-Evid) specifically designed to exploit these diagnostic findings.
    \item We conduct an extensive 18-model audit demonstrating that legacy single-attack proxies produce unstable rankings for native models. Our diagnostic-driven attacks substantially raise the estimated residual visual-channel risk, showing that current evaluations can underestimate VLM vulnerabilities.
\end{itemize}

\section{Related Work}
\label{sec:related}

\paragraph{Native Vision-Language Models.}
Early Vision-language Models (VLMs) typically used modular architectures~\cite{clip, dai2023instructblip, yang2023dawn}, connecting a frozen or lightly adapted vision encoder to a language model through a projection layer, query module, or connector~\cite{alayrac2022flamingo, xu2023mplug, wang2024cogvlm}. Models like BLIP-2~\cite{li2023blip2}, InstructBLIP~\cite{dai2023instructblip}, MiniGPT-4~\cite{zhu2023minigpt4}, and LLaVA~\cite{liu2023llava} showed that this alignment-era pipeline supports visual question answering, captioning, and visual instruction tuning~\cite{yue2024mmmu, fu2024mme}. Later work systematically examined how vision backbones, connectors, and training recipes affect VLM behavior~\cite{karamcheti2024prismatic, tong2024cambrian, chen2024internvl, lu2024deepseek, qwen3vl}. Recent native multimodal models instead adopt multi-layer visual injection or early-fusion shared backbones, allowing visual and textual information to interact earlier or deeper in computation~\cite{team2024chameleon, yu2023cm3leon, chern2024anole, lu2024unifiedio2, zhan2024anygpt, qwen3_5, gemma4_2026}. This shift means the visual channel is no longer merely a frontend conditioning signal, so robustness evaluation must move beyond assumptions inherited from modular pipelines.

\paragraph{Adversarial Robustness of VLMs.}
Classical adversarial robustness work shows that small-norm pixel perturbations and PGD-style attacks can expose systematic vulnerabilities in visual models~\cite{carlini2017towards, madry2018towards}. These methods have been adapted to VLMs to attack vision encoder representations, cross-modal alignment, or final generated responses~\cite{schlarmann2023adversarial, tu2023many, zhao2024evaluating, zhang2022coattack}. Prior studies further show that large VLMs remain sensitive to visual perturbations and that adversarial examples can transfer across models~\cite{schlarmann2024robust, carlini2024aligned, zhang2025veattack, xie2025coa, wang2025doublevisual, wang2026omni, zhang2024advpt, zhang2025cavalry, zhang2025anyattack, zhang2026reasonbreak, zhang2026naptuning}. Beyond pixel-space attacks, typographic attacks demonstrate that text rendered inside images can alter a VLM's interpretation of visual content~\cite{goh2021multimodal, materzynska2022disentangling, gong2025figstep, cheng2025tvpi}. However, existing evaluations usually treat pixel perturbations and text overlays as separate threat vectors, without systematically measuring joint risk across visual pathways in native VLMs~\cite{bailey2023imagehijacks, bagdasaryan2023abusing, greshake2023not, qi2024visual, shayegani2024jailbreak}.

\paragraph{Robustness Evaluation and Diagnostic Auditing.}
Robustness evaluation has long recognized that single attacks can underestimate risk~\cite{athalye2018obfuscated}, motivating fixed attack suites such as AutoAttack~\cite{croce2020reliable}. Multimodal robustness evaluations have begun to aggregate diverse perturbations and attacks, but most still rely on predefined static suites~\cite{tu2023many}. In parallel, diagnostic tools including representation similarity, attention analysis, and activation patching help reveal internal representations and behavioral mechanisms~\cite{basu2024understanding, tong2024eyes, golovanevsky2025notice, neo2025towards, li2025diffuguard}. Yet these diagnostics are typically used post hoc to explain failures rather than to guide stronger subsequent attacks~\cite{kornblith2019similarity, meng2022locating, geva2023dissecting}. PaRMA instead \mbox{integrates} pathway-specific diagnostics into the audit, reporting residual risk after diagnostic-driven strengthening rather than the maximum over a pre-specified collection.

\section{Native Multimodal VLMs: Taxonomy and Threat Scope}
\label{sec:native_scope}

The transition from modular pipelines to native multimodal training alters the VLM attack surface. In pipeline models, a projection bottleneck isolates the vision encoder from the language backbone. Deep fusion removes this bottleneck, altering end-to-end gradient paths, while joint early-layer attention forces visual and textual tokens to compete directly. To systematically audit this evolving landscape, we introduce a graded nativeness taxonomy and formalize the visual threat scope.

\subsection{Nativeness Taxonomy}
\label{sec:taxonomy}

We categorize VLMs based on their fusion depth and pretraining strategy. Fusion level co-varies with model family, scale, and training data, so level-wise trends are correlational. This taxonomy serves as an organizational axis for our audit, rather than a strict causal ladder:

\begin{itemize}[leftmargin=*, nosep]
    \item \textbf{\levelOne{} (Alignment-era):} A frozen or lightly adapted vision encoder connects to a language model via a projection module~\cite{liu2023llava, zhu2023minigpt4}. Modalities remain in structurally separate spaces; the bottleneck is the only junction.
    \item \textbf{\levelTwo{} (Shallow fusion):} Models feature large-scale joint multimodal pretraining. Visual tokens enter the backbone through a connector, but do not re-enter computation at deeper layers; the backbone processes a fixed visual prefix.
    \item \textbf{\levelThree{} (Deep/early fusion):} Visual information participates throughout the language computation. This includes DeepStack (DS) models that re-inject visual features across multiple layers~\cite{qwen3vl}, and Early Fusion (EF) models that share transformer blocks between modalities from the first layer~\cite{qwen3_5, gemma4_2026}.
\end{itemize}
A theoretical fourth level (\levelFour{})---a strict single-backbone design unifying all modalities under one tokenizer without a distinct vision tower---completes the continuum, though no audited open-weight model currently meets this threshold.

\subsection{Audit Scope and Visual Pathways}
\label{sec:visual_pathways}

We formalize a VLM as a function mapping a visual input $x_v$ and a text prompt $x_t$ to an answer. To isolate visual-channel vulnerabilities from text-only jailbreaks, our audit fixes the text prompt and intervenes exclusively on $x_v$. Crucially, all empirical Attack Success Rates (ASRs) in our evaluation are strictly conditioned on the model's \textit{clean-correct set}---the subset of examples the model answers correctly before any intervention.

We decompose the visual attack surface into two primary pathways:
\begin{enumerate}[leftmargin=*, nosep]
    \item \textbf{Perceptual corruption ($\pi_{\text{pix}}$):} Pixel-space perturbations. In deep-fusion architectures, the unbroken gradient path from output loss to pixels can expose failures missed by attacks centered on \levelOne{}-style bottlenecks.
    \item \textbf{Rendered-text insertion ($\pi_{\text{txt}}$):} Typographic overlays. Added image text competes with existing visual evidence and the user's prompt within the joint attention mechanism.
\end{enumerate}

\textbf{The Audit Problem.} Given a model and a dataset, a \textit{pathway-complete robustness audit} estimates the failure probability over the clean-correct set induced by the union of pathway-specific interventions. Crucially, it should diagnose whether these failures exhibit behavioral signatures that can guide stronger attacks.

\section{Method: PaRMA and FusionAudit}
\label{sec:parma}

\begin{figure}[t]
  \centering
  \includegraphics[width=\textwidth]{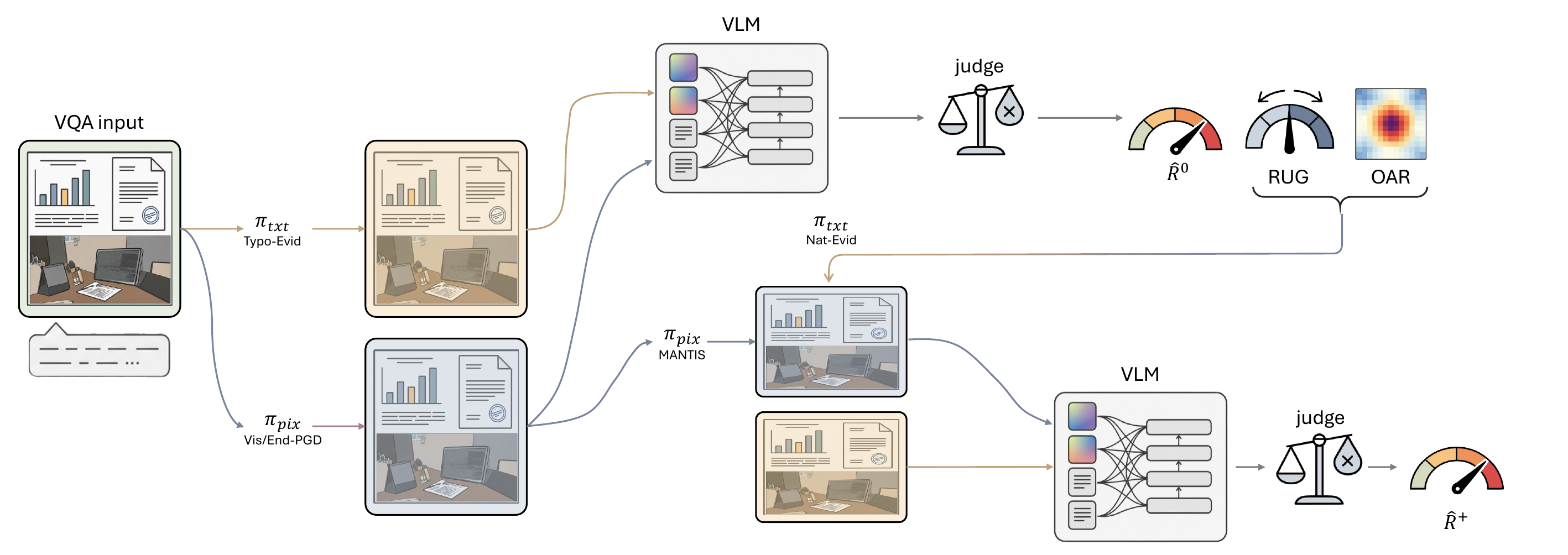}
  \caption{\textbf{The PaRMA framework and FusionAudit pipeline.} Native VLMs are evaluated over two visual pathways: pixel perturbation and rendered-text insertion. Baseline attacks (Vis-PGD, End-PGD, Typo-Evid) yield initial failures and diagnostic readouts. These diagnostics surface architecture-associated failure signatures (e.g., the \textit{see-but-not-deceived} signature), guiding the generation of diagnostic-motivated augmented attacks (MANTIS, Nat-Evid) to compute the final falsifiable residual risk $\widehat{R}^+$.}
  \label{fig:framework}
\end{figure}

We operationalize our threat scope through \textbf{PaRMA} (Pathway-conditioned Risk for Multimodal Auditing). Unlike static benchmarks, PaRMA is a diagnostic-driven audit protocol: baseline attacks yield diagnostic signals, which in turn guide the design of diagnostic-motivated augmented attacks. We instantiate this framework as \textbf{FusionAudit} over two visual pathways (Figure~\ref{fig:framework}).

\subsection{Pathway-Conditioned Risk Formulation}
\label{sec:formulation}
\label{sec:risk}

We treat the visual input $x_v$ as a site for pathway-specific interventions. A pathway $k$ is an intervention family $(\Bk,\phik)$, where $\Bk$ is the perturbation domain and $\phik(x_v;\delta)$ is the intervened image. FusionAudit instantiates two pathways:
\begin{align*}
\pi_{\text{pix}} &: \Bpix = \{\delta:\|\delta\|_\infty\le \varepsilon\}, \quad \phipix(x_v;\delta)=\operatorname{clip}(x_v+\delta), \\
\pi_{\text{txt}} &: \Btxt = \Sigma^{\le L}\times\mathcal{P}, \quad \phitxt(x_v;s,p)=\operatorname{render}(x_v;s,p).
\end{align*}
Here $\Sigma^{\le L}$ is the space of candidate rendered strings of length at most $L$, and $\mathcal{P}$ denotes admissible render placements; $s$ is the inserted string and $p$ its placement.
Given a finite attack suite $\Aset_k$ for pathway $k$, let $\Set{a} = \{z\in\mathcal{C}_\theta(\mathcal{D}): \Judge(f_\theta(a(x_v),x_t),y)=0\}$ be the subset of clean-correct examples broken by attack $a$. The empirical pathway-complete risk is the union failure rate:
\begin{equation}
\label{eq:fa_estimator}
\Rhat(f_\theta;\mathcal{D},\Aset) = \frac{\left|\bigcup_{k}\bigcup_{a\in\Aset_k}\Set{a}\right|}{|\mathcal{C}_\theta(\mathcal{D})|}.
\end{equation}
PaRMA's departure from naive multi-attack unions is that $\Rhat$ is not static; it is a lower bound strengthened via pathway-resolved diagnostics.

\subsection{Baselines and Diagnostic Suite}
\label{sec:baselines_diagnostics}

The baseline protocol assigns standard attacks to each pathway to probe primary failure modes.

For perceptual corruption ($\pi_{\text{pix}}$), we employ two PGD variants bounded by $\|\delta\|_\infty\le\varepsilon$. Let $h_\mathcal{V}(x_v)$ denote the flattened vision-encoder output. \textbf{Vis-PGD} anchors the loss at the vision encoder, minimizing cosine similarity to the clean feature map:
\begin{equation}
\label{eq:vispgd}
\min_{\|\delta\|_\infty\le\varepsilon}\cos\!\bigl(h_\mathcal{V}(\phipix(x_v;\delta)),\,h_\mathcal{V}(x_v)\bigr).
\end{equation}
\textbf{End-PGD} anchors the loss at the end-to-end output, maximizing cross-entropy over the full teacher-forced correct answer sequence $y^+$:
\begin{equation}
\label{eq:endpgd}
\max_{\|\delta\|_\infty\le\varepsilon}\Bigl[-\!\sum_{t=1}^{|y^+|}\log p_\theta\!\bigl(y^+_t\mid\phipix(x_v;\delta),x_t,y^+_{<t}\bigr)\Bigr].
\end{equation}

For rendered-text insertion ($\pi_{\text{txt}}$), following \citet{cheng2024typod}, \textbf{Typo-Evid} superimposes a task-matched wrong-answer string $w$ at a fixed region $p$ as a direct typographic overlay:
\begin{equation}
\label{eq:typoe}
\tilde{x}_v = \phitxt(x_v;\,w,\,p).
\end{equation}

To analyze rendered-text baseline failures, PaRMA uses two diagnostics.
\textbf{Read--Use Gap (RUG).}
For $\pi_{\text{txt}}$, let $E_{\mathrm{read}}(z)$ denote the event that the rendered wrong-answer string $w$ is recovered under an explicit reading prompt, and let $E_{\mathrm{use}}(z)$ denote the event that the task-prompt response is wrong and contains $w$. RUG is therefore a prompt-conditioned accessibility--adoption diagnostic. Over the clean-correct set,
\begin{align}
P_{\mathrm{read}}(f_\theta)
  &= \Pr_{z\sim\mathcal{D}}\!\left[E_{\mathrm{read}}(z)\mid z\in\mathcal{C}_\theta(\mathcal{D})\right], \notag\\
P_{\mathrm{use}}(f_\theta)
  &= \Pr_{z\sim\mathcal{D}}\!\left[E_{\mathrm{use}}(z)\mid z\in\mathcal{C}_\theta(\mathcal{D})\right], \notag\\
\mathrm{RUG}(f_\theta)
  &= P_{\mathrm{read}}(f_\theta)-P_{\mathrm{use}}(f_\theta).
\label{eq:rug}
\end{align}
\textbf{Overlay Attention Ratio (OAR).}
For the Typo-Evid image, let $\Omega \subset [N_v]$ be the overlay's image-patch index set, where $N_v$ is the number of image patches. Let $\bar{a}_j$ denote the mean attention weight from the text tokens after the image to image patch $j$, averaged over heads and late decoder layers. OAR quantifies the fraction of image-directed attention falling on the overlay relative to its area:
\begin{equation}
\label{eq:oar}
\mathrm{OAR} =
\frac{\sum_{j\in\Omega}\bar{a}_j / \sum_{j=1}^{N_v}\bar{a}_j}{|\Omega|/N_v}.
\end{equation}
\textbf{Diagnostic Signature.} Together, high task-prompt OAR ($>1$) and positive RUG are consistent with a \textit{see-but-not-deceived} signature in native VLMs: the overlay is accessible under explicit reading, receives disproportionate task-prompt attention, but is not reliably adopted in the task response. This suggests a visual-distinctiveness artifact in the Typo-Evid baseline.

\subsection{Diagnostic-Driven Augmentation}
\label{sec:diagattacks}

The baseline analyses indicate two limitations: End-PGD spreads gradient pressure across the sequence (Eq.~\ref{eq:endpgd}), and Typo-Evid triggers visual rejection. We introduce two diagnostic-motivated augmentations to close these gaps.

\textbf{MANTIS} (\underline{M}argin-\underline{A}ware \underline{N}ative \underline{T}oken \underline{I}ntervention \underline{S}trategy). To strengthen the pixel pathway audit, MANTIS abandons full-sequence loss. Let $y^+$ be the canonical correct continuation and $y^-$ be a deterministic canonical wrong continuation derived from the reference answer. Both continuations are fixed before optimization. MANTIS concentrates gradient pressure exclusively at $t^\star$, the first token where the two continuations diverge:
\[
t^\star = \min\{t : y^+_t \neq y^-_t\}.
\]
It then optimizes the margin at this first-divergence position:
\begin{equation}
\label{eq:mantis}
\max_{\|\delta\|_\infty\le\varepsilon} \Bigl[ \log p_\theta(y^-_{t^\star}\mid\phipix(x_v;\delta),x_t,y^-_{<t^\star}) - \log p_\theta(y^+_{t^\star}\mid\phipix(x_v;\delta),x_t,y^+_{<t^\star}) \Bigr].
\end{equation}

\textbf{Nat-Evid} (Naturalized Evidence). To test whether host-congruent rendering reduces this read-without-use gap, Nat-Evid replaces the typographic overlay with a style-conditioned image edit $\hat{x}_v = \mathcal{E}(x_v;\,w)$. The wrong answer $w$ is rendered as host-congruent native content (e.g., a chart annotation or scene text) rather than an extrinsic superimposition. The wrong answer $w$ is fixed before generation, and the editor is instructed to add text only, without modifying or removing existing content.

\textbf{Strengthened Estimator.} By augmenting the attack families with MANTIS and Nat-Evid, we compute the strengthened risk $\RhatPlus$. The gap $\RhatPlus-\RhatZero$ measures the additional risk surfaced by our diagnostics. $\RhatPlus$ serves as the audit's principal, falsifiable deliverable.

\section{Experiments}
\label{sec:experiments}

\subsection{Experimental Setup}
\label{sec:setup}

\textbf{Datasets and Models.} We evaluate on a fixed 1500-sample VQA evaluation set, balanced across ChartQA~\cite{masry2022chartqa}, DocVQA~\cite{mathew2021docvqa}, and MME~\cite{fu2024mme}. These datasets span document understanding and visual reasoning scenarios where images naturally contain both perceptual content and rendered text. ChartQA and DocVQA are text-heavy, whereas 460 of the 500 MME examples (92\%) come from natural-image categories (400 artwork, 60 celebrity) and the remaining 40 from a text-dense category (code reasoning). Restricting the MME evaluation to these natural images changes no level-mean ASR by more than 3.2 points, and the \levelThree{} MANTIS exclusive-flip rate is $37.0\%$, against $36.4\%$ on all 500 MME examples, so the level-wise patterns also hold on natural-image VQA. We select open-weight VLMs that cover the proposed nativeness levels while keeping parameter scales within the same broad range whenever possible, so that level comparisons are not dominated by a small-versus-large model contrast. The final 18-model inventory contains four \levelOne{} models (SmolVLM2-2.2B, BLIP2-OPT-2.7B, InstructBLIP-Vicuna-7B, LLaVA-1.5-7B), seven \levelTwo{} models (Qwen-VL-Chat, Qwen2-VL-2B, Qwen2.5-VL-3B/7B, Gemma3-4B/12B, DeepSeek-VL-7B), and seven \levelThree{} models (Qwen3-VL-2B/4B/8B, Qwen3.5-2B/4B/9B, Gemma4-E4B). Within this inventory, Qwen2.5-VL, Qwen3-VL, and Qwen3.5 form a same-family controlled chain from shallow to deep fusion at nearby parameter scales, which helps contextualize the cross-family level aggregates.

\textbf{Evaluation Protocol.} For paired comparison, all attacks are evaluated on the exact same 1500-sample evaluation set. Crucially, all Attack Success Rates (ASRs) and risk estimators are conditioned strictly on each model's \textit{clean-correct set} ($n_c$)---the subset of examples the model answers correctly before any intervention. Pixel attacks use an $\ell_\infty$ budget of $\varepsilon=8/255$ with 40 PGD steps. When a model exhausts memory at the native image size, we compute the perturbation on a resized image, upsample only the noise field, and add it back to the original-resolution image before inference.

\subsection{Main Results}
\label{sec:main_audit}

Table~\ref{tab:main_audit_new} presents the main pathway-complete audit across the 18 models. We report Attack Success Rates (ASR) for the baseline suite ($\Aset^0$: Vis-PGD, End-PGD, Typo-Evid) and the diagnostic-driven strengthened suite ($\Aset^+$: MANTIS, Nat-Evid). The principal reported scalar is the pathway-complete risk estimator $\widehat{R}^+$, representing the union of failures across all strengthened visual pathways.

\begin{table*}[!t]
\centering
\caption{\textbf{Pathway-Complete Robustness Audit.} Results are aggregated over the three-dataset audit set (ChartQA, DocVQA, and MME). All ASRs are percentages conditioned on each model's clean-correct set ($n_c$). Gray parentheticals show the absolute ASR gain over the paired baseline (e.g., MANTIS vs.\ End-PGD). Crucially, the $\delta$ columns report the \textit{exclusive-flip rate}: the fraction of $n_c$ broken by the augmented attack that survived \textit{all} corresponding baseline attacks (e.g., MANTIS $\setminus$ (Vis-PGD $\cup$ End-PGD)). \auditbglabel{Light gray} columns mark the final audit-summary risks. Level means are unweighted macro-averages.}
\label{tab:main_audit_new}
\scriptsize
\setlength{\tabcolsep}{0pt}
\renewcommand{\arraystretch}{1.10}
\begin{threeparttable}
\begin{tabular*}{\textwidth}{@{\extracolsep{\fill}} l
  S[table-format=2.1]
  S[table-format=2.1] S[table-format=2.1] c S[table-format=2.1]
  S[table-format=2.1] c S[table-format=2.1]
  >{\columncolor{AuditSummaryBg}[2pt][2pt]}S[table-format=3.1]
  >{\columncolor{AuditSummaryBg}[2pt][2pt]}c @{}}
\toprule
\multicolumn{2}{c}{} & \multicolumn{4}{c}{Pixel pathway} & \multicolumn{3}{c}{Rendered-text pathway} & \multicolumn{2}{c}{Audit summary} \\
\cmidrule(lr){3-6}\cmidrule(lr){7-9}\cmidrule(l){10-11}
Model & {Clean} & {Vis-PGD} & {End-PGD} & {MANTIS} & {$\delta$} & {Typo-Evid} & {Nat-Evid} & {$\delta$} & {$\widehat{R}^{0}$} & {$\widehat{R}^{+}$} \\
\midrule
\multicolumn{11}{@{}l}{\textit{\levelOne{}: alignment-only / projection-based models}} \\
SmolVLM2-2.2B          & 61.2 & 52.7 & 80.4 & $99.5$\dpos{19.1} & 13.4 & 17.2 & $30.9$\dpos{13.7} & 22.2 & 86.7 & $100.0$\dpos{13.3} \\
BLIP2-OPT-2.7B         & 19.8 & 20.2 & 43.1 & $63.0$\dpos{19.9} & 25.9 & 69.4 & $4.4$\dneg{65.0}  &  1.0 & 83.5 & $91.2$\dposp{7.7}  \\
InstructBLIP-Vicuna-7B & 26.5 & 60.5 & 70.5 & $98.0$\dpos{27.5} & 11.8 & 10.8 & $6.0$\dnegp{4.8}  &  3.0 & 87.7 & $99.5$\dpos{11.8}  \\
LLaVA-1.5-7B           & 30.7 & 47.5 & 82.0 & $94.8$\dpos{12.8} &  9.5 &  8.5 & $20.0$\dpos{11.5} & 15.2 & 87.6 & $97.2$\dposp{9.6}  \\
\addlinespace[1pt]
\textit{\levelOne{} mean (4)} & 34.5 & 45.2 & 69.0 & $88.8$\dpos{19.8} & 15.2 & 26.5 & $15.3$\dneg{11.2} & 10.4 & 86.4 & $97.0$\dpos{10.6} \\
\addlinespace[2pt]
\midrule
\multicolumn{11}{@{}l}{\textit{\levelTwo{}: shallow native multimodal fusion}} \\
Qwen-VL-Chat           & 57.5 & 71.8 & 62.2 & $93.3$\dpos{31.1} & 13.5 & 23.7 & $34.8$\dpos{11.1} & 22.9 & 85.3 & $97.9$\dpos{12.6}  \\
Qwen2-VL-2B            & 72.4 & 17.8 & 65.9 & $66.9$\dposp{1.0}  & 16.7 & 31.3 & $38.3$\dposp{7.0}  & 22.7 & 74.5 & $89.8$\dpos{15.3}  \\
Qwen2.5-VL-3B          & 77.3 & 19.4 & 66.6 & $59.5$\dnegp{7.1}  & 13.0 & 19.5 & $36.6$\dpos{17.1} & 26.8 & 72.0 & $85.7$\dpos{13.7}  \\
Gemma3-4B              & 65.3 & 45.2 & 45.3 & $71.5$\dpos{26.2} & 18.9 & 11.1 & $33.8$\dpos{22.7} & 27.2 & 66.5 & $85.4$\dpos{18.9}  \\
Qwen2.5-VL-7B          & 80.1 & 13.3 & 45.4 & $59.0$\dpos{13.6} & 25.2 & 15.6 & $33.9$\dpos{18.3} & 24.3 & 54.2 & $79.5$\dpos{25.3}  \\
Gemma3-12B             & 73.6 & 38.1 & 35.7 & $69.5$\dpos{33.8} & 22.5 & 14.1 & $34.0$\dpos{19.9} & 25.1 & 60.2 & $81.5$\dpos{21.3}  \\
DeepSeek-VL-7B         & 52.9 & 79.2 & 58.0 & $96.3$\dpos{38.3} & 13.5 & 31.1 & $29.9$\dnegp{1.2}  & 16.8 & 87.9 & $99.2$\dpos{11.3}  \\
\addlinespace[1pt]
\textit{\levelTwo{} mean (7)} & 68.4 & 40.7 & 54.2 & $73.7$\dpos{19.5} & 17.6 & 20.9 & $34.5$\dpos{13.6} & 23.7 & 71.5 & $88.4$\dpos{16.9} \\
\addlinespace[2pt]
\midrule
\multicolumn{11}{@{}l}{\textit{\levelThree{}: deep-stack or early-fusion native models}} \\
Qwen3-VL-2B            & 78.0 & 11.8 & 42.8 & $62.9$\dpos{20.1} & 29.1 & 10.4 & $32.1$\dpos{21.7} & 27.1 & 51.1 & $82.7$\dpos{31.6}  \\
Qwen3-VL-4B            & 80.7 & 11.8 & 37.4 & $61.1$\dpos{23.7} & 32.1 & 12.6 & $29.9$\dpos{17.3} & 23.5 & 44.7 & $78.9$\dpos{34.2}  \\
Qwen3-VL-8B            & 82.5 &  9.0 & 41.6 & $56.6$\dpos{15.0} & 25.5 &  4.0 & $27.8$\dpos{23.8} & 24.8 & 46.0 & $73.1$\dpos{27.1}  \\
Qwen3.5-2B             & 77.7 &  9.4 & 39.9 & $61.9$\dpos{22.0} & 31.3 & 13.5 & $31.2$\dpos{17.7} & 22.1 & 46.5 & $78.0$\dpos{31.5}  \\
Qwen3.5-4B             & 79.5 & 12.1 & 31.7 & $43.9$\dpos{12.2} & 22.8 & 27.2 & $24.5$\dnegp{2.7}  & 10.0 & 50.4 & $68.1$\dpos{17.7}  \\
Qwen3.5-9B             & 82.7 & 11.1 & 38.3 & $39.2$\dposp{0.9}  & 14.3 & 16.1 & $25.1$\dposp{9.0}  & 14.5 & 43.4 & $59.5$\dpos{16.1}  \\
Gemma4-E4B             & 62.7 & 50.1 & 46.7 & $72.4$\dpos{25.7} & 17.3 &  8.9 & $27.1$\dpos{18.2} & 22.2 & 71.4 & $88.7$\dpos{17.3}  \\
\addlinespace[1pt]
\textit{\levelThree{} mean (7)} & 77.7 & 16.5 & 39.8 & $56.8$\dpos{17.0} & 24.6 & 13.2 & $28.2$\dpos{15.0} & 20.6 & 50.5 & $75.6$\dpos{25.1} \\
\bottomrule
\end{tabular*}
\end{threeparttable}
\end{table*}

\textbf{The Diagnostic-Driven Risk Gap.}
The main finding is that the diagnostic-driven suite yields higher risk estimates than the baseline suite. Across all fusion levels, the diagnostic-driven attacks expose additional failures. Moving from $\widehat{R}^0$ to $\widehat{R}^+$, the level-mean risk increases by $+10.6\%$ in \levelOne{}, $+16.9\%$ in \levelTwo{}, and $+25.1\%$ in \levelThree{}. Fixed-model paired bootstraps put the corresponding 95\% CIs at $[9.2,12.0]$, $[16.1,17.8]$, and $[24.2,26.0]$ percentage points, all excluding zero.

The $\delta$ columns and the $\widehat{R}^{+}$ parentheticals make the additional-risk contribution of diagnostic-driven augmentation explicit. For the pixel pathway, the fraction of clean-correct examples broken exclusively by MANTIS---examples that survived \textit{both} Vis-PGD and End-PGD---rises from $15.2\%$ in \levelOne{}, to $17.6\%$ in \levelTwo{}, and peaks at $24.6\%$ in \levelThree{}. This supports our diagnostic premise: in deeper-fusion architectures, standard end-to-end loss can spread gradient pressure across the sequence, leaving failures that are surfaced by concentrating gradient at the answer-discriminative token. On the rendered-text pathway, Nat-Evid surfaces large but non-monotone exclusive-flip rates ($10.4\%$, $23.7\%$, and $20.6\%$ across \levelOne{}--\levelThree{}), with a particularly large $20.6\%$ of \levelThree{} clean-correct examples flipped after resisting Typo-Evid. At the full audit level, the exclusive additional-risk fraction grows from $10.6\%$ to $16.9\%$ to $25.1\%$, even while the absolute residual risk ($\widehat{R}^+$) decreases in newer models ($97.0\% \to 88.4\% \to 75.6\%$).

\textbf{Within-Family Qwen Chain.}
The Qwen series provides a crucial within-family control to mitigate confounders---such as parameter scale and training data evolution---across our \levelOne{}--\levelThree{} taxonomy. Around the smaller scale, $\widehat{R}^{+}$ decreases from Qwen2.5-VL-3B (\levelTwo{}, 85.7\%) to Qwen3-VL-2B/4B (\levelThree{}, 82.7/78.9\%) and Qwen3.5-2B/4B (\levelThree{}, 78.0/68.1\%). Around the larger scale, it decreases from Qwen2.5-VL-7B (\levelTwo{}, 79.5\%) to Qwen3-VL-8B (\levelThree{}, 73.1\%) and Qwen3.5-9B (\levelThree{}, 59.5\%). This same-family pattern is consistent with the overall level-wise trend in Table~\ref{tab:main_audit_new}. Crucially, while absolute vulnerability ($\widehat{R}^{+}$) is lower in newer models, the \textit{diagnostic-driven risk gap} ($\widehat{R}^{+}-\widehat{R}^{0}$) persists across the chain. This pattern is consistent with the view that MANTIS and Nat-Evid expose vulnerabilities associated with the native multimodal shift, rather than only reflecting a comparison between older and newer model families.

\textbf{Ranking Instability and the Limits of Single-Attack Proxies.}
Historically, End-PGD ASR has served as the default proxy for VLM robustness. Our audit shows that this proxy can be insufficient for native architectures. Comparing the End-PGD ranking to the pathway-complete $\widehat{R}^+$ ranking yields a Spearman $\rho=0.837$ and Kendall $\tau_b=0.634$---correlated, but unstable enough to affect model-selection conclusions. Specifically, 8 out of 18 models shift by at least three ranks. For instance, Gemma3-12B appears highly robust under End-PGD (ranking 2nd), but its higher MANTIS ASR moves it to 7th under $\widehat{R}^+$. This instability supports the motivation for PaRMA: pathway-specific failures can be missed by single-scalar, pixel-centric evaluations.

\subsection{Pixel Pathway Deep Dive}
\label{sec:pixel_deep_dive}

\begin{figure*}[!ht]
  \centering
  \begin{subfigure}[b]{0.5\textwidth}
      \centering
      \includegraphics[width=\textwidth]{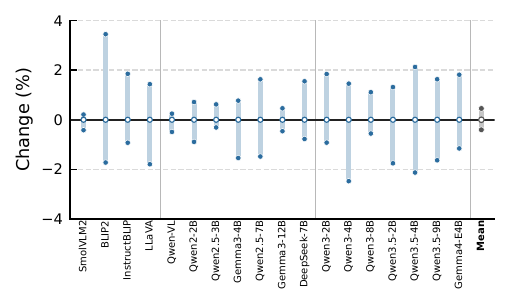}
      \caption{Random-start sensitivity.}
      \label{fig:stab_seed}
  \end{subfigure}
  \hfill
  \begin{subfigure}[b]{0.99\textwidth}
      \centering
      \includegraphics[width=\textwidth]{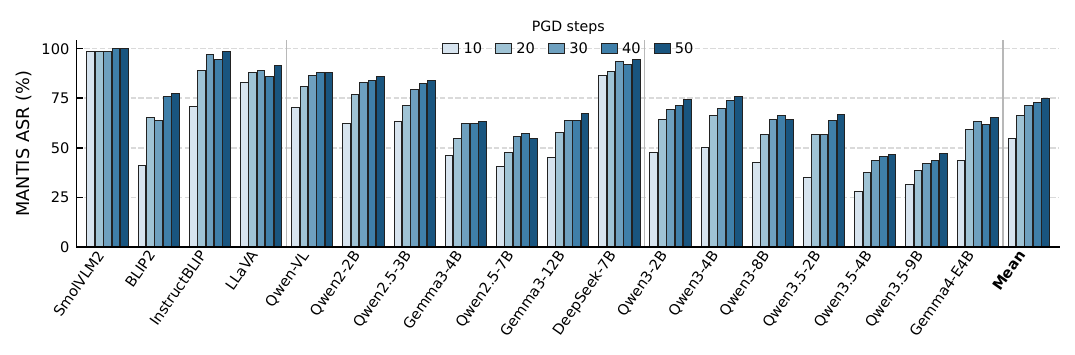}
      \caption{Optimization steps ($t$).}
      \label{fig:stab_steps}
  \end{subfigure}
  \hfill
  \begin{subfigure}[b]{0.99\textwidth}
      \centering
      \includegraphics[width=\textwidth]{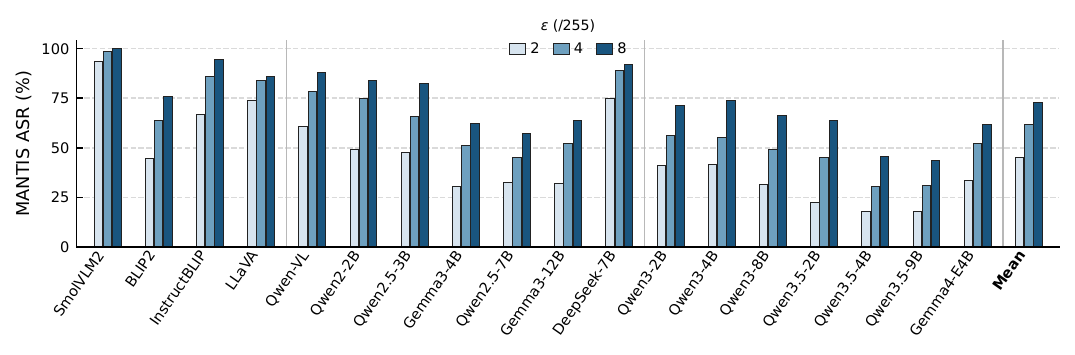}
      \caption{Perturbation budget ($\varepsilon$).}
      \label{fig:stab_eps}
  \end{subfigure}
  \caption{\textbf{MANTIS Stability and Ablation.} Evaluated on a balanced 300-sample subset across all 18 models. (a) Min-max ASR variance across three seeds, centered on the per-model mean. (b) Clean-correct ASR as PGD steps increase at $\varepsilon=8/255$. (c) Clean-correct ASR across varying $\ell_\infty$ radii at $t=40$. The rightmost bars in each panel show the 18-model mean.}
  \label{fig:mantis_stability}
\end{figure*}

\textbf{Ablation and Stability.} One concern for white-box attacks is whether observed gains are artifacts of specific hyperparameter tuning. We conduct a stability ablation on a balanced 300-sample subset (100 from each dataset) across all 18 models. Figure~\ref{fig:mantis_stability} suggests that MANTIS's gains are not only an artifact of hyperparameter tuning:

\textit{Random-Start Sensitivity (Fig.~\ref{fig:stab_seed}):} We evaluate MANTIS across three seeds. The min-max variance is small relative to the absolute ASR, with a median standard deviation of just $1.3\%$. This suggests that MANTIS is not heavily dependent on initialization.

\textit{Optimization Budget (Fig.~\ref{fig:stab_steps}):} We sweep the PGD step count from 10 to 50 to verify that our 40-step audit budget is not under-optimized. The results show that ASR largely saturates by 30--40 steps. Pushing the budget to 50 steps yields only marginal level-mean gains ($+2.8\%$, $+1.0\%$, and $+2.2\%$ for Levels 1--3), indicating that the 40-step setting is a reasonable audit budget.

\textit{Perturbation Radius (Fig.~\ref{fig:stab_eps}):} We vary the $\ell_\infty$ radius $\varepsilon \in \{2, 4, 8\}/255$. While the absolute ASR naturally scales with the perturbation budget, the relative vulnerability ordering across fusion levels (\levelOne{} $>$ \levelTwo{} $>$ \levelThree{}) is preserved in the tested settings. This suggests that the trend exposed by MANTIS is not specific to the chosen threat model capacity.

\textbf{Layer Restoration.} To locate where End-PGD flips are decided, we restore clean decoder activations on End-PGD-flipped examples. For each model, we take the first 40 clean-correct examples flipped by End-PGD (sorted by identifier), regenerate their adversarial images with the audit configuration, and keep the flips that reproduce (24--40 per model). For each depth $\ell$, we replace the prefill hidden states of decoder layers $0$ through $\ell$ with their values on the clean image, decode greedily, and count the flip as recovered if some depth restores the correct answer. Recovery falls with fusion level: $92.2\%$ $[87.6, 96.1]$ in \levelOne{}, $62.6\%$ $[55.5, 69.6]$ in \levelTwo{}, and $25.6\%$ $[20.7, 30.7]$ in \levelThree{} (model-macro means with 95\% fixed-model bootstrap CIs over 16 models; Qwen-VL-Chat and DeepSeek-VL-7B are excluded because their legacy implementations do not support activation hooks). Within the Qwen family, it falls from $70.4\%$ (Qwen2-VL-2B, Qwen2.5-VL-3B/7B) to $21.3\%$ (Qwen3-VL and Qwen3.5). In alignment-only models, restoring an early prefix of layers usually undoes the attack; in deep-fusion models, most flips survive restoration at every depth, consistent with visual features re-entering the computation across many layers. This is where an answer-level objective that targets the decision token, as in MANTIS, adds the most exclusive flips (Table~\ref{tab:main_audit_new}).

\subsection{Text Pathway Deep Dive}
\label{sec:text_deep_dive}

On the rendered-text pathway, Table~\ref{tab:main_audit_new} separates direct typographic overlays from host-congruent rendered text. Typo-Evid is strongest in \levelOne{}, driven by one alignment-era model, whereas Nat-Evid is stronger than Typo-Evid by $13.6\%$ in \levelTwo{} and $15.0\%$ in \levelThree{}. This pattern suggests that direct overlays are not a uniform proxy for rendered-text risk in native models: the payload may be processed, but not necessarily adopted as the task answer.

\begin{table}[t]
\centering
\caption{\textbf{Read--Use Gap.} Values are unweighted level means over the 18-model audit.}
\label{tab:rug_level}
\scriptsize
\setlength{\tabcolsep}{4pt}
\begin{tabular}{@{}lccc@{}}
\toprule
Level & {$P_{\mathrm{read}}$} & {$P_{\mathrm{use}}$} & {RUG} \\
\midrule
\levelOne{} & 38.9 & 22.4 & +16.5 \\
\levelTwo{} & 68.6 & 15.9 & +52.7 \\
\levelThree{} & 81.0 & 11.6 & +69.5 \\
\bottomrule
\end{tabular}
\end{table}

\textbf{Behavioral Evidence (RUG).} Table~\ref{tab:rug_level} reports the corresponding behavioral diagnostic from Eq.~\ref{eq:rug}. The alignment-only group is heterogeneous, but the native groups show a clear level-wise trend: $P_{\mathrm{read}}$ increases from \levelTwo{} to \levelThree{} while $P_{\mathrm{use}}$ decreases, widening RUG from $+52.7\%$ to $+69.5\%$. Because $P_{\mathrm{read}}$ is measured with an explicit reading prompt, we interpret RUG as accessibility-versus-adoption evidence; together with OAR, it is consistent with the \textit{see-but-not-deceived} interpretation and motivates Nat-Evid as a host-congruent wrong-answer payload.

\begin{figure*}[!ht]
  \centering
  \begin{minipage}[c]{0.55\textwidth}
    \includegraphics[width=\textwidth]{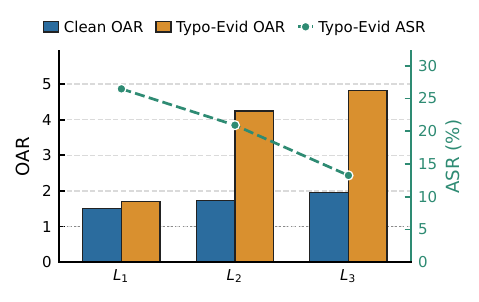}
  \end{minipage}\hfill
  \begin{minipage}[c]{0.42\textwidth}
    \includegraphics[width=\textwidth]{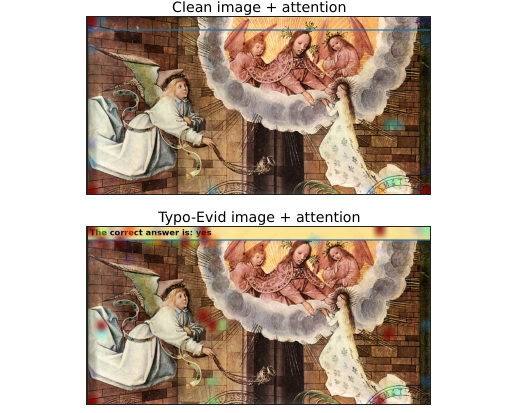}
  \end{minipage}
  \caption{\textbf{Attention-Level Evidence for the \textit{See-but-not-Deceived} Signature.} \textbf{(Left)} Level-mean Overlay Attention Ratio (OAR). In \levelTwo{} and \levelThree{}, OAR more than doubles over clean ($+138\%$ and $+135\%$), while Typo-Evid ASR falls with fusion level. \textbf{(Right)} Qualitative attention map for Qwen3.5-2B on an MME sample. Attention concentrates on the injected banner (blue box), while the model keeps the visual answer and outputs "no".}
  \label{fig:oar_analysis}
\end{figure*}

\begin{figure}[t]
  \centering
  \includegraphics[width=0.44\textwidth]{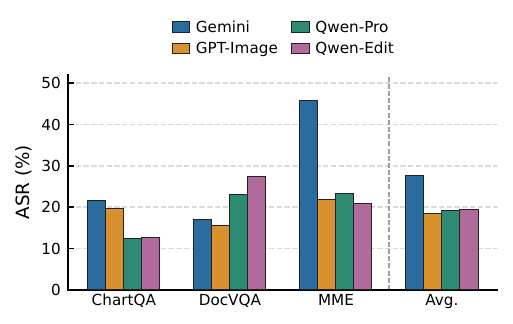}
  \caption{\textbf{Generator Sensitivity of Nat-Evid.} Model-macro ASR across four generated variants.}
  \label{fig:generator_comparison}
\end{figure}

\textbf{Attention-Level Evidence (OAR).} We next ask whether the read--use gap has a corresponding attention-level signature. We compute the Overlay Attention Ratio (OAR, Eq.~\ref{eq:oar}) against the matched clean image to isolate the attention shift caused by the attack.

Figure~\ref{fig:oar_analysis} (Left) provides the aggregate attention evidence. The mean OAR increase is $+138\%$ in \levelTwo{} and $+135\%$ in \levelThree{}, against $+13\%$ in \levelOne{}. From \levelTwo{} to \levelThree{}, Typo-Evid ASR drops from $20.9\%$ to $13.2\%$. In other words, native models allocate far more attention to the overlay, while deeper-fusion models less often adopt it as the answer.

Figure~\ref{fig:oar_analysis} (Right) illustrates this pattern. On a clean-correct MME sample, Qwen3.5-2B is attacked with a typographic banner asserting "The correct answer is: yes". The answer-query attention map concentrates on the injected banner (OAR rises from 2.49 to 4.85), but the model still outputs the correct visual answer ("no"). This attention-behavior gap is consistent with the Nat-Evid augmentation: to affect a native VLM's text pathway, the payload may need to be not only readable but visually congruent enough to be adopted.

\textbf{Generator Sensitivity.} Because Nat-Evid relies on generative image editing, we test whether the attack's success is tied to the idiosyncrasies of a specific generator. Figure~\ref{fig:generator_comparison} compares four Nat-Evid variants. While the Gemini-backed variant is the strongest (driven by MME), the three exact-text generators (GPT-Image-2 and two Qwen variants) produce highly consistent average ASRs ($\sim 19\%$). This suggests that the Nat-Evid effect is not tied to a single generator.

\textbf{Ground-Truth Preservation.} Nat-Evid edits are generated under strict text-only constraints, and we verify that they keep the original evidence intact. For natural images, we mask the inserted cue with a solid black patch on 100 MME examples that Qwen3-VL-8B answers correctly on the original image but wrongly under Nat-Evid; the correct answer returns in 98 of the 100 cases, so these failures come from the injected cue rather than from altered evidence. For charts and documents, a non-generative OCR engine recovers the exact ground-truth string from the edited image in 92.3\% of the examples whose answer it can read in the original, and manual inspection of the remaining cases finds the evidence destroyed in only 0.85\% of the images.

\section{Conclusion and Limitations}
\label{sec:conclusion}
\label{sec:limitations}

The transition to native multimodal architectures fundamentally alters the VLM attack surface, rendering legacy single-attack, pixel-centric evaluations insufficient. To address this, we introduced \textbf{PaRMA} and its instantiation, \textbf{FusionAudit}, an evidence-based auditing methodology. By leveraging pathway-specific diagnostics, we identified an objective mismatch on the pixel pathway and the \textit{see-but-not-deceived} signature in deep-fusion models. These insights motivated our diagnostic-motivated interventions, MANTIS and Nat-Evid, which successfully exposed substantial residual risks systematically hidden by standard benchmarks across 18 VLMs. Ultimately, robustness auditing must evolve alongside architectures; FusionAudit provides a rigorous, pathway-complete template for this native multimodal era.

PaRMA can help model developers find visual-channel failures before deployment. The same attacks could also be used to mislead deployed VLMs; MANTIS requires white-box gradient access, and we release the audit pipeline rather than precomputed adversarial images.

Our study has several limitations. First, our nativeness taxonomy (\levelOne{}--\levelThree{}) is an observational framework for auditing, not a strict causal ladder. The stricter \levelFour{} endpoint remains outside our empirical audit: token-based mixed-modal systems such as Chameleon point toward this design, but the Chameleon paper reports image-tokenizer limitations on text-heavy images, which are central to our ChartQA, DocVQA, and rendered-text setting~\cite{team2024chameleon}. While the Qwen controlled chain mitigates some confounders, adjacent levels differ in training data and scale alongside fusion depth. Second, our rendered-text pathway focuses exclusively on task-relevant wrong-answer injection. Command-following visual prompt injection (e.g., jailbreaks) represents a distinct safety risk that requires compliance-based metrics rather than correctness-based ASR. Finally, while PaRMA currently instantiates two visual pathways, the framework is designed to be extensible. As VLMs expand to process audio, video, and embodied tool-use, new pathways and diagnostics can be added to PaRMA.

\begin{ack}
This research is also supported by the Ministry of Education, Singapore, under its Academic Research Fund Tier 2 (Award MOE-T2EP20125-0005).
\end{ack}

\bibliographystyle{plainnat}
\bibliography{refs}

\end{document}